\documentclass[letterpaper, 10 pt, conference]{ieeeconf}  

\IEEEoverridecommandlockouts    

\usepackage{amsmath,amssymb,amsfonts}
\usepackage{graphicx}
\usepackage{epstopdf,epsfig}
\usepackage{color}
\usepackage{algorithm,algorithmic,subcaption}
\usepackage{multirow}
\usepackage{hyperref}
\usepackage[colorinlistoftodos]{todonotes}
\usepackage[utf8]{inputenc}
\usepackage{balance}

\graphicspath{{./figures/}}

\newtheorem{theorem}{Theorem}[section]

\newtheorem{problem}[theorem]{Problem}
\newtheorem{definition}{Definition}
\newtheorem{remark}{Remark}

\newcommand{\wpt}[2]{p^{(#1)}_{#2}}
\newcommand{\wptt}[2]{t^{(#1)}_{#2}}
\newcommand{\wptpair}[2]{\left(\wpt{#1}{#2},\wptt{#1}{#2}\right)}
\newcommand{\wptsec}[3]{T_{#1}\left(#2\,.\,.\,#3\right)}
\newcommand{\wptime}[1]{\mathtt{time}\left(#1\right)}
\newcommand{\wppos}[1]{\mathtt{pos}\left(#1\right)}

\begin{document}

\title{\LARGE \bf Autonomous Planning for Multiple Aerial Cinematographers}

\author{
Luis-Evaristo Caraballo$^{1}$, Ángel Montes-Romero$^{2}$, José-Miguel Díaz-Báñez$^{1}$, Jesús Capitán$^{2}$,   \\
Arturo Torres-González$^{2}$
 and Aníbal Ollero$^{2}$
\thanks{$^{1}$Department of Applied Mathematics II, University of Seville, Seville, Spain}
\thanks{$^{2}$GRVC Robotics Lab, University of Seville, Seville, Spain}
}

\maketitle
\thispagestyle{empty}
\pagestyle{empty}

\begin{abstract}
This paper proposes a planning algorithm for autonomous media production with multiple \emph{Unmanned Aerial Vehicles} (UAVs) in outdoor events. Given \emph{filming tasks} specified by a media \emph{Director}, we formulate an optimization problem to maximize the filming time considering battery constraints. As we conjecture that the problem is NP-hard, we consider a discretization version, and propose a graph-based algorithm that can find an optimal solution of the discrete problem for a single UAV in polynomial time. Then, a greedy strategy is applied to solve the problem sequentially for multiple UAVs. We demonstrate that our algorithm is efficient for small teams (3-5 UAVs) and that its performance is close to the optimum. We showcase our system in field experiments carrying out actual media production in an outdoor scenario with multiple UAVs.
\end{abstract}


\section{Introduction}

The use of \emph{Unmanned Aerial Vehicles} (UAVs) for cinematography and audio-visual applications is becoming quite trendy. First, small UAVs are not expensive and can be equipped with high-quality cameras that are available in the market for amateur and professional users. Second, they can produce unique video streams thanks to their maneuverability and their advantageous viewpoints when flying. 


Currently, these systems use two operators per aerial camera at least: a pilot for the vehicle and a camera operator. Our final objective is to develop a team of cooperative UAV cinematographers with autonomous functionalities. This would reduce the number of human operators and would allow the media \emph{Director} to focus on artistic aspects. This Director is the person in charge of the media production, specifying desired shots to cover a certain event. Then, the multi-UAV system should be able to plan and execute autonomously the designed shots. This planning problem is challenging in large-scale scenarios, due to multiple constraints: (i) spatial constraints related to no-fly areas; (ii) temporal constraints regarding the events to film; and (iii) resource constraints related to the battery endurance of each UAV. 

\begin{figure}[htb]
	\centering
	\includegraphics[width=\linewidth]{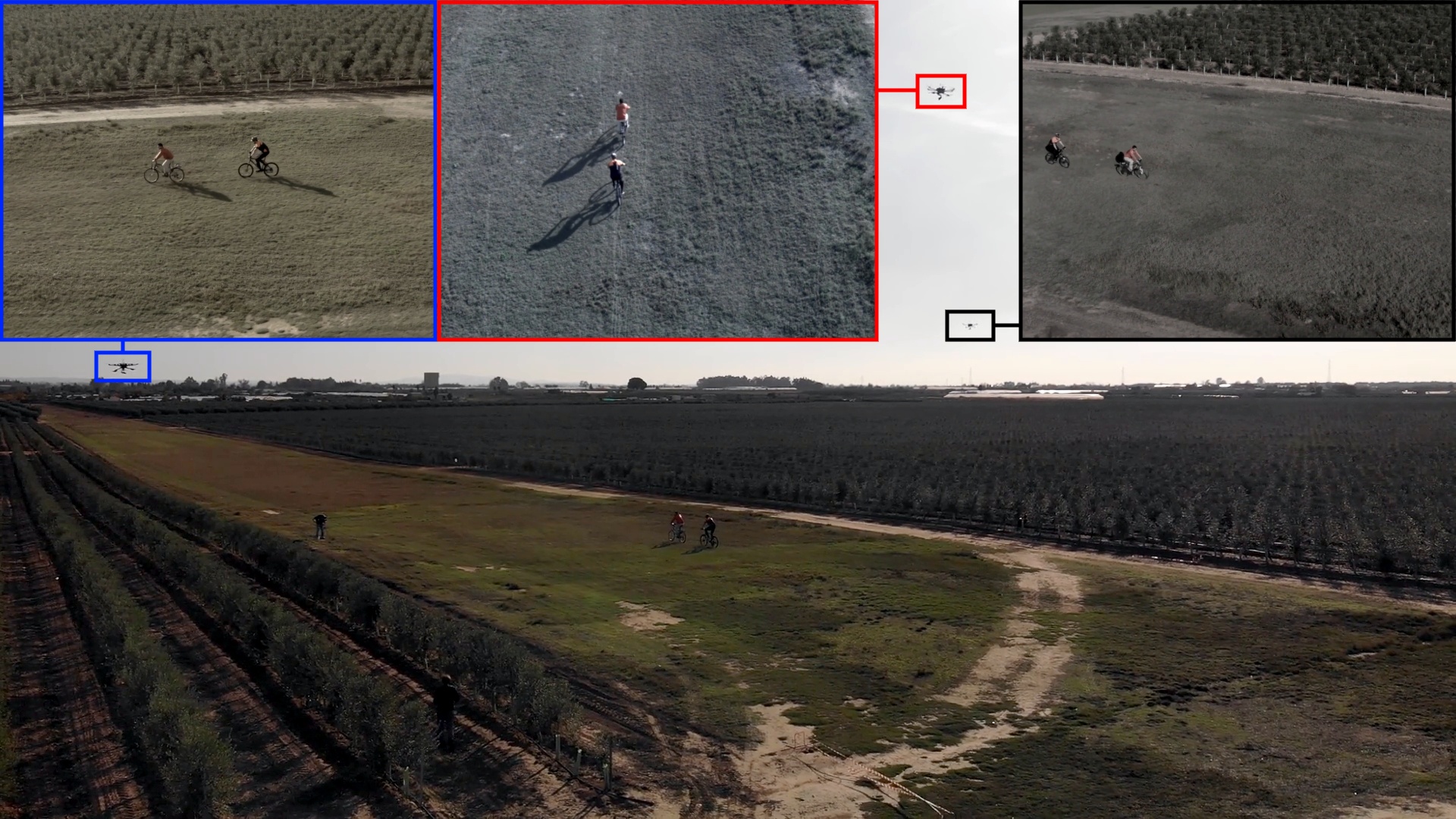}
	\caption{Experimental media production with three UAVs taking different shots in parallel. Bottom, aerial view of the experiment with two moving cyclists. Top, images taken from the cameras on board each UAV. }
	\label{fig:MissionOf3Drones}
\end{figure}

\setcounter{footnote}{2}

This paper proposes a multi-UAV approach for autonomous cinematography planning, aimed at filming outdoor events such as cycling or boat races (see Fig.~\ref{fig:MissionOf3Drones}). The work lies within the framework of the EU-funded project MULTIDRONE~\footnote{https://multidrone.eu}, which has developed autonomous media production with small teams of UAVs. In previous work~\cite{montes_appsci20}, we proposed a graphical interface and a novel language so that the media production Director could design \emph{shooting missions}. Thus, the Director specifies the characteristics of multiple shots that should be executed to film a given event. We call these shots \emph{filming tasks} or \emph{shooting actions}, and they are described basically by their starting time and duration, the starting camera position and relative positioning with respect to the target, and the geometrical formation of the cameras depending on the type of shot.  

In this paper, we use the Director's shooting mission as input and propose an algorithm to compute plans for the UAVs. Given their bounded battery, the problem to schedule UAVs to cover the filming tasks is a complex optimization problem hard to solve. After surveying related works (Section~\ref{sec:relatedWork}), we contribute by formulating this novel optimization problem for autonomous filming (Section~\ref{sec:problem}). Then, we propose a graph-based solution to the problem (Section~\ref{sec:discretization}) that can find an optimal solution for a single UAV and approximate solutions for the multi-UAV case. 


Our algorithm is deterministic and assumes that trajectories of all moving targets in the event can be predicted with a given motion model. However, in order to account for model uncertainties and unexpected events (e.g., a UAV failure), we enhance the planner with a module that monitors the mission execution and triggers re-planning to address deviation from the original plan. Re-planning could also be applied at a given rate, in order to account for updates in the models to predict target motion. Thus, plans are computed online in a centralized manner, which is feasible for small teams (3-5 UAVs).  
Moreover, we evaluate the performance of our algorithm through simulation (Section~\ref{sec:simulations}) and showcase its integration for actual media production with multiple UAVs (Section~\ref{sec:experiments}).




\section{Related Work}
\label{sec:relatedWork}
There are several commercial products, such as DJI GO, AirDog, 3DR SOLO or Yuneec Typhoon, that implement \emph{follow-me} autonomous capabilities to track a target visually or with a GPS. These approaches do not consider high-level cinematography principles for shot performance and just try to keep the target on the image. Additionally, there are some recent works to produce semi-autonomous aerial cinematographers~\cite{joubert_arxiv16,gebhardt_chi16}. In these works, the user or Director specifies high-level commands such as shot types and positions, and the drone is in charge of implementing autonomously the navigation functionality. In \cite{joubert_arxiv16}, an outdoor application to film people is proposed, and different types of shots from the cinematography literature are introduced (e.g., close-up, external, over-the-shoulder, etc). 
In \cite{gebhardt_chi16}, an iterative quadratic optimization problem is formulated to obtain smooth trajectories for the camera and the look-at point (i.e., place where the camera is pointing at). No time constraints nor moving targets are included. \cite{bonatti_arxiv19} propose an autonomous system for outdoor, aerial cinematography, but they do not explore the multi-UAV problem.

Some works~\cite{naegeli_tg17} propose camera motion planning to achieve smooth trajectories, but in this paper, we focus on high-level planning. This means how to distribute filming tasks among the team members. As aerial shots can be viewed as tasks to be executed by the UAVs, algorithms for multi-robot task allocation~\cite{Nunes2017} are of interest. In~\cite{B.KartalE.Nunes2016}, a centralized algorithm based on a Monte-Carlo Tree Search is presented for multi-robot task allocation. The algorithm exploits the \emph{branch\&bound} paradigm to solve the problem. Others~\cite{Jones201141} have proposed both centralized and decentralized methods to deal with disaster management. 

 
For aerial cinematography, it is also relevant to consider formulations where there are time constraints associated with the tasks~\cite{Nunes2017}, as visual shots may be triggered and executed with specific timelines. An auction-based method is presented in~\cite{Nunes2017a}. They decouple precedence and temporal constraints and treat them separately with two different algorithms that can work offline (producing a schedule of tasks in advance) and online (scheduling tasks as they arrive). \cite{Luo2015876} present a distributed algorithm for multi-robot task assignment where the tasks have to be completed within given deadlines, taking into account the battery constraints. For the special case of constant task duration they present a distributed algorithm that is provably almost optimal.



Recently, a similar optimization problem to that stated in this paper,
has been addressed for static targets and continuous time intervals~\cite{austriacos}. They prove that the problem with unlimited battery lifetime can be solved in polynomial time by  finding a maximum weight matching in a weighted bipartite graph. They conjecture that the optimization problem with limited battery lifetime is NP-hard and thus, the same conjecture remains for moving targets.

\section{Problem Statement}
\label{sec:problem}
In this section, we formally define the \emph{Cycling Filming Problem} (CFP).
Suppose that an outdoor event is to be filmed by a set of $k$ UAVs with cameras and limited battery endurance given by a parameter $b$.
The media Director specifies a set ${\cal T}=\{T_1,\dots,T_n\}$ of shooting actions (tasks) determined by waypoints and time intervals during which the UAVs should film the moving targets (e.g., cyclists). That is, a shooting action $T_i\in \cal{T}$, that starts at time $t$ and ends at time $t'>t$, is determined by a list $(p_1, t_1),\dots, (p_s, t_s)$ of pairs, where $t=t_1<\dots<t_s=t'$ and $p_j$ is the filming position of the camera at time $t_j$~\footnote{These filming positions are computed after predicting the target trajectory and depending on the shot type.}.
A task, 
or part of it, can be performed by one or multiple UAVs (e.g., a UAV may film the first part of a task and the rest may be filmed by a different one). A \emph{flight plan} for a UAV is a sequence $P=\{I_1,\dots,I_m\}$ such that for every $j$, $I_j$ is a subinterval of a task in $\cal{T}$. 
Denote by ${\cal P}=\{P_1,\dots,P_k\}$ the set of the flight plans for the $k$ UAVs.
The goal is to assign flight plans
to each UAV in order to film as much as possible of the set ${\cal T}$. The \emph{filming time}
of a flight plan assignment ${\cal P}$ is the sum of the time lengths of the subintervals of the tasks covered by the flight plans. Formally, it is defined
as $$FT({\mathcal P})=\sum_{i=1}^{n}\left |\bigcup_{
I\in \,\cup P_j
} \left(I \cap T_i \right )\right |.$$

There is one or more \emph{Base Stations} (BS) that can be static or dynamic, where the UAVs start from and where they can go back to recharge their battery at any time. It is assumed that it is possible to compute a path between any pair of locations. Also, having this path, it is assumed that the required time to travel along the path can be estimated as well as the expected cost in terms of battery. Thus, it can be checked at any moment whether a UAV has enough battery to return to a BS. 

	\begin{problem}(CFP)\label{prob:maxtime}
	Given a set ${\cal T}$ of $n$ tasks and $k$ UAVs, each with battery endurance $b$, find a flight plan assignment maximizing the filming time.
	\end{problem}
	
\begin{remark}
 For simplicity of presentation, we have assumed that charging at BS occurs instantaneously. However, our results could be easily extended considering a time $\delta$ to recharge the battery when a UAV arrives at a BS.
\end{remark}



Notice that, in a solution, a UAV can enter into a task or it can leave it at any instant of the task's time interval. Hence, CFP is in general a continuous optimization problem that is conjectured to be NP-hard~\cite{austriacos}.


\section{The Approximation Algorithm}
\label{sec:discretization}
In this section, we consider a discretization based on time of the general problem that allows us to obtain an approximation to the CFP in polynomial time.
Our discretization is based on the construction of a directed acyclic graph $G$ whose vertices are pairs $(p, t)$, where $p$ is a position and $t$ is an associated instant of time. The discretization is carried out based on time instead of distance for convenience with the application, as we try to maximize the filming time, i.e., the number of discrete pieces of time covered in total. In the following we briefly show how to build this graph.

\begin{figure}
    \centering
    \includegraphics[width=.8\columnwidth, page=2]{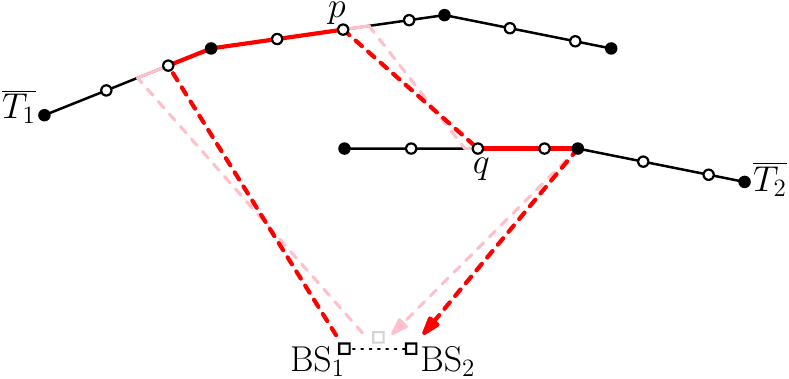}
    \caption{The hollow points represent the introduced/added waypoints in the tasks in the discretization process. The red (dark) path is an approximation using the discretization waypoints of the light (pink) path. BS$_1$ and BS$_2$ refer to a moving Base Station in two different instants of time.
    }
    \label{fig:round-discretization}
\end{figure}

Let $\alpha<<b$ be a real positive value. For every two consecutive waypoints of a task, subdivide the task into pieces with time length $\alpha$ except for maybe the last piece. Let $\overline{T_i}$ denote the augmented task adding the new waypoints of the partition as illustrated in Fig.~\ref{fig:round-discretization}. 

The waypoints of the tasks $\overline{T_i}$, $i=1,\cdots,n$ are  vertices of the graph $G$.
For every two consecutive pairs $(p_j, t_j)$ and $(p_{j+1},t_{j+1})$ in a task $\overline{T_i}$ add the edge $\left((p_j,t_j),(p_{j+1},t_{j+1})\right)$ to $G$. We say that a waypoint $(p',t')$ is \emph{reachable} from a waypoint $(p, t)$ if $t'-t$ is greater than or equal to the required time to travel from $p$ to $p'$. Now, given two tasks, $\overline{T_i}$ and $\overline{T_i'}$, connect every pair $(p,t)$ in $\overline{T_i}$ with the first waypoint $(p',t')\in \overline{T_{i'}}$ which is \emph{reachable} from $(p,t)$. In Fig.~\ref{fig:round-discretization}, $q$ is the first waypoint of $\overline{T_2}$ which is reachable from $p$. 

Let us add now the vertices and edges related to the BS. For every task vertex $(p,t)$ and every BS $\beta$ add the vertex $(p',t')$ and the edge $((p',t'),(p,t))$ where $p'$ and $t'$ are the departure position and departure time to leave BS $\beta$ and to arrive at position $p$ exactly at time $t$. Analogously, add the vertex $(p'',t'')$ and the edge $((p,t), (p'',t''))$ where $(p'', t'')$ denotes the arriving waypoint at BS $\beta$ departing from position $p$ at time $t$. Finally, add edges between consecutive waypoints of a same BS. And, if there are more than one BS, for every waypoint $(p,t)$ in a BS $\beta$, add an edge from $(p,t)$ to $(p',t')$ where $(p',t')$ is the first reachable waypoint of another BS $\beta'$.

Notice that, for every vertex $(p,t)$ in $G$, its outgoing edges goes toward vertices $(p',t')$ such that $t'>t$, then the graph $G$ is acyclic. Also, notice that every edge $e$ has an associated travel time which is the time difference between the connected vertices. If the edge connects consecutive vertices of a same task then this time difference is also the filming time value of the edge, $FT(e)$. In other case, the filming time is zero.


Let $P=[(p_1,t_1),\dots,(p_m,t_m)]$ be a path in $G$ such that $(p_1,t_1)$ and $(p_m,t_m)$ are BS vertices. $P$ corresponds to a flight plan if for every BS vertex  $(p_i,t_i)$ in $P$ ($i<m$) the next BS vertex $(p_j, t_j)$ fulfills that $t_j-t_i\leq b$ or, in other case, $j=i+1$ and both vertices correspond to the same BS.

An approximate solution for the CFP is to find $k$ of such paths in $G$ maximizing the sum of the filming time values of the traversed edges.

\begin{problem}(DCFP)
Given a discretization graph $G$ and a battery capacity $b$, computes $k$ paths (flight plans) in $G$ such that the filming time is maximized. That is:
$$\text{Maximize\quad}\sum_{e\in\displaystyle\cup_{i=1}^k P_i}FT(e),$$
 where $P_i$ denotes the set of edges of the path (flight plan) assigned to the $i$-th UAV.

\end{problem}

\subsection{Greedy strategy}
\label{sec:method}

In order to alleviate the complexity of the multi-UAV optimization problem, we propose a greedy strategy. This means that we first solve the problem considering a single UAV; then, we remove the sub-intervals covered by this UAV from the tasks, and solve the remaining tasks with another UAV. Applying this approach sequentially, we can obtain an approximation to the optimal solution with $k$ UAVs.

\subsubsection{Algorithm for one UAV}
Consider that the battery consumption is constant in time (normally is not, but, the middle rate of battery consumption can be used). Then,
if the UAV is at vertex $u$ with battery level $l$, it can traverse the edge $(u,v)$ only in the following cases:
\begin{enumerate}
    \item $u$ and $v$ are both vertices of the same base station (the UAV stays at the BS and it is not consuming energy).
    \item $u=(p,t)$, $v=(p',t')$ and $l \geq t'-t+r$, being $r$ the battery capacity needed to return back to the closest base station from $p'$.
\end{enumerate}

We can prove the following result:

\begin{theorem}
Given a discretization graph $G$ with parameter $\alpha$, and a battery capacity $b$, the problem DCFP for one UAV can be solved in $O\left(b\left(\frac{mn}{\alpha}\right)^2\sum_{i=1}^n|T_i|\right)$ time by using dynamic programming, where $n$ and $m$ are  the  number  of filming tasks and  Base Stations, respectively.
\end{theorem}

\begin{remark}
Notice that the smaller the $\alpha$ value is, the more accurate and time consuming the algorithm is.
\end{remark}

\subsubsection{Strategy for a small team of UAVs}

For the case in which a small team of UAVs is used, the one-UAV algorithm can be applied iteratively, updating the graph by removing the visited arcs and removing the assigned UAVs. 
The order in which UAVs are selected to run their task assignment is done in an optimal manner, by taking at each iteration the one that maximizes its filming time.
Of course, once the paths are computed, the UAVs would work in parallel, assuming that an avoiding collision approach is integrated into the vehicles.
Although our greedy strategy is not optimal in general, it has linear scalability with the number of UAVs. The algorithm for $k$ UAVs has complexity:
\begin{equation}\label{eq:k_drones}
O\left(kb\left(\frac{mn}{\alpha}\right)^2\sum_{i=1}^n|T_i|\right)\text{ time}.
\end{equation}

\section{Simulation Experiments}
\label{sec:simulations}


This section depicts some experiments to assess the performance of our algorithm for DCFP. Several tests were performed to carry out two studies: (i) to validate the quality of the method; and (ii) to justify that a small team of UAVs is enough to guarantee optimal filming time in certain scenarios.  

Notice that the filming time depends on the parallelization of the tasks and the lengths of the associated time intervals. Obviously, if at some instant $t$ there are $x$ active tasks 
and $k<x$ then it is not possible to cover all the proposed filming tasks. And even, if at any time, the number of active tasks is equal to or less than $k$, it may still not be possible to cover all of them due to battery constraints. 

In order to measure the quality of the proposed method, we use the \emph{coverage ratio}, which is the filming time using $k$ UAVs over the sum of the time lengths of all tasks, that is:
$$CR=\displaystyle\frac{\sum^k_{i=1}\sum_{e\in P_i}FT(e)}{\sum_{i=1}^n\left|T_i\right|}.$$

\subsubsection{Experiment 1. Coverage ratio vs number of UAVs}
Given two natural numbers $n$ and $x$, a scenario $A$ is generated randomly with $n$ tasks distributed longitudinally along a route such that at any time there are at most $x$ active tasks. The parameters chosen randomly for each longitudinal task are the starting and ending positions and time, such that $x$ is always fulfilled. After that, the developed algorithm is run on $A$ incorporating UAVs one by one until all the tasks are covered.



\begin{figure}[th]
    \centering
    \includegraphics[width=\linewidth]{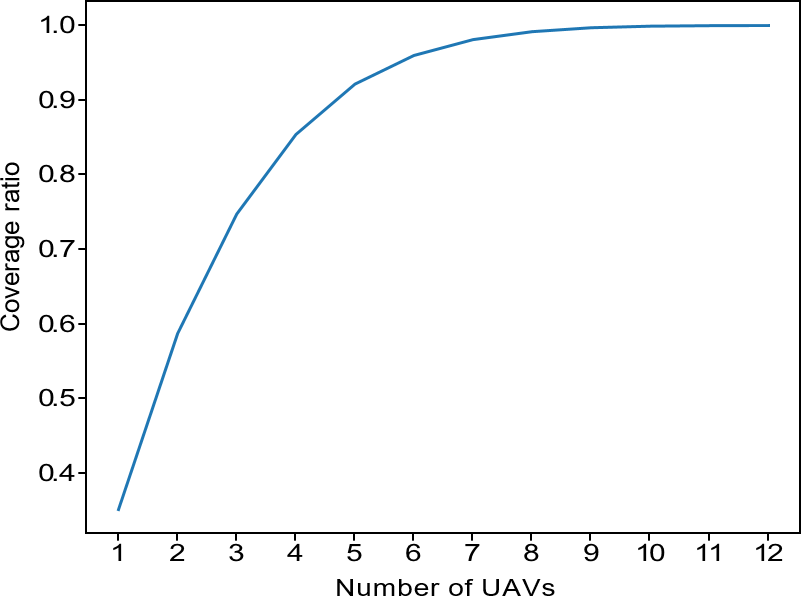}
    \caption{Coverage ratio vs number of UAVs using the average data obtained from 50 repetitions of the experiment using $n=20$ and $x=4$. 
    }
    \label{fig:cumulative-20-4}
\end{figure}

Random scenarios with $n=10, 15, 20, 25$ and $x=4,6,8$ were generated and
we repeated the experiment 50 times for every pair of values $n$ and $x$.
Figure~\ref{fig:cumulative-20-4} shows the average results for $n=20$ and $x=4$. For any average case ($n,x$) a similar behavior was obtained: using $k$ UAVs, with $k$ around $x$, a coverage ratio around 0.8 is achieved and, for $k>x$, the greater $k$ is, the lower the increase of the coverage ratio is. 
Moreover, although we do not know the optimal value, $\mathcal{O}$, of the coverage ratio using $k\geq x$ drones, the experiment shows that our result, $\mathcal{R}$, meets that $0.8\leq \mathcal{R}\leq \mathcal{O}\leq 1$, that is, the approximation factor $\frac{\mathcal{O}}{\mathcal{R}}$ is between 1 and 1.25. We experimentally study this factor with an additional experiment.

\subsubsection{Experiment 2. Approximation factor}
We compare the accuracy of the proposed greedy approach against an exhaustive method over the graph $G$.
For that, we generated simulation scenarios choosing parameters randomly from realistic value ranges: targets move following a straight line with a variable speed of 1-2m/s; the maximum speed of the UAVs is 3m/s; a full battery is considered to last 15 minutes; shooting actions have a length of at most 80m and their time-lengths are between 30 and 70 seconds~\footnote{This values are recommendations from the media experts in MULTIDRONE project in line with actual media productions.}. With this range of parameters, we generated randomly four types of shooting actions: (1) \emph{Static}, a single point where the UAV stays to film a target or panoramic views during a time interval; (2) \emph{Chase}, a path parallel to the to target trajectory using the same speed of the target to follow it from behind during this time interval; (3) \emph{Flyby}, a path parallel to the target trajectory but using greater speed in order to film the target from back to front; and (4) \emph{Orbit}, a circular arc that crosses the target trajectory from one side to another.

\begin{figure}
    \centering
    \includegraphics[width=\columnwidth]{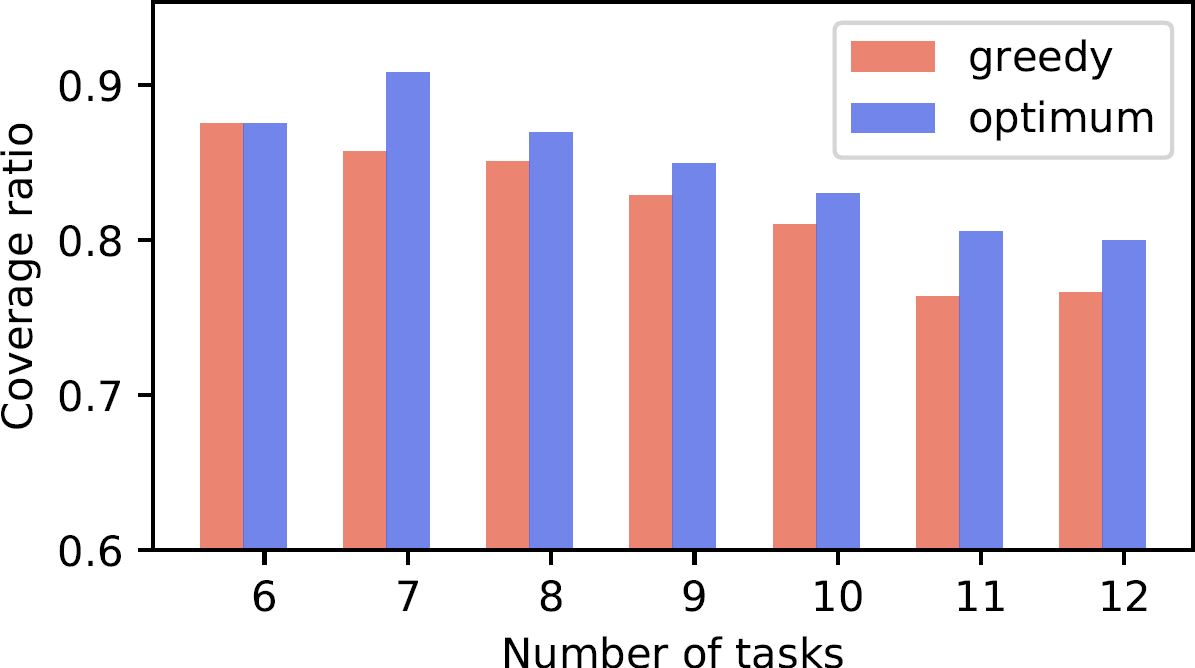}
    \caption{Optimal versus approximated $CR$.}
    \label{fig:comparing}
\end{figure}

Due to the huge complexity of the exhaustive approach we used $\alpha=30$s in order to get small discretization graphs. We used $k=3$ UAVs and we limited the temporal overlapping of the tasks to three, that is, there are always at most 3 active tasks. We generated several experiments and grouped the results by the number of generated tasks. Then, we computed the average of the coverage ratio in each group for the two strategies: our \emph{greedy} proposal and the \emph{optimal} exhaustive analysis. The results are shown in Fig.~\ref{fig:comparing}. Notice that our proposal has a very good performance and is much more efficient than the exhaustive method, which requires exponential time.

\section{Field Experiments}
\label{sec:experiments}

In this section, we describe the integration of our algorithm with a real team of UAVs performing autonomous media production.

\subsection{System integration}

\begin{figure}[htb]
	\centering
	\includegraphics[width=\linewidth]{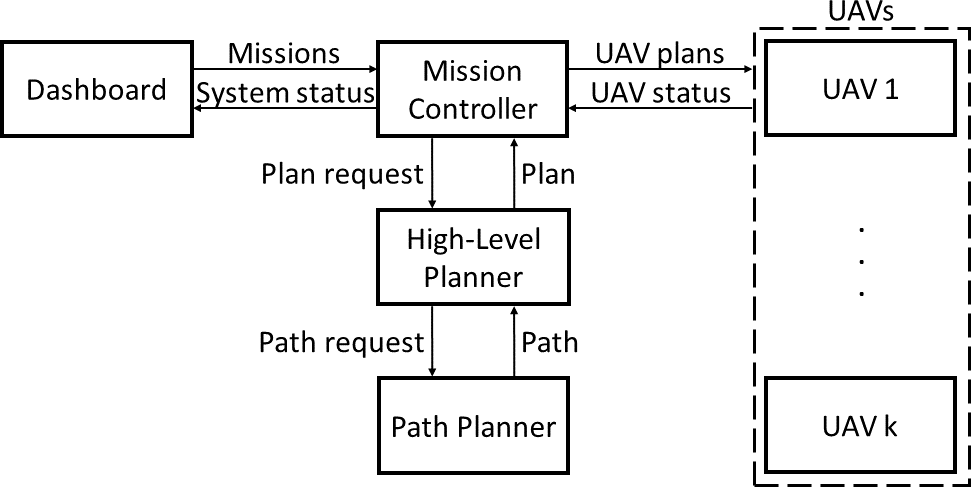}
	\caption{Block diagram of our system architecture. The planning process occurs in a centralized fashion at a ground station.}
	\label{fig:ArchitectureSimplified}
\end{figure}

Figure~\ref{fig:ArchitectureSimplified} depicts the block diagram of our software architecture. All modules have been implemented in C++ and integrated with ROS Kinetic Kame~\footnote{Code online at \href{https://bitbucket.org/multidrone_eu/multidrone_full}{https://bitbucket.org/multidrone\_eu/multidrone\_full.}}. Our planning algorithm for shooting missions is the core of the system and runs on the \emph{High-Level Planner} (HLP). The workflow of the system starts with the Director describing the shooting mission through the \emph{Dashboard}, which is an intuitive GUI. We described the Dashboard and the process to transform shooting missions into filming tasks in previous work~\cite{montes_appsci20}, where we proposed a novel language for the description of media missions. The \emph{Mission Controller} (MC) receives the mission with the shooting actions and requests the HLP for a plan. Once the plan is computed, the MC sends to each UAV its part of the plan. Then, those associated filming tasks are executed on board each UAV in a distributed fashion. This procedure for distributed mission execution is out of the scope of this paper and more details can be seen in~\cite{alcantara_arxiv20}. The MC is also in charge of monitoring continuously the status of the UAVs and asking for re-planning. Thus, if a UAV has an emergency (e.g., it runs out of battery) and has to stop its current action, the MC would request the HLP for a new plan with the current initial conditions (battery and positions) of the available UAVs and the remaining filming tasks. This re-planning could also be applied periodically at a given rate, in order to account for updates in the prediction models for target motion and battery consumption.

The \emph{Path Planner} is an auxiliary module so that the HLP can compute collision-free paths (and associated costs) between waypoints. As we endeavor to produce media in outdoor areas, no-fly zones for the UAVs are also taken into account (e.g., to avoid the area with the audience). We implemented an A* algorithm for path planning in a map grid containing known no-fly zones and obstacles. Note that local collision avoidance would still be needed during mission execution. 

\begin{figure}[htb]
	\centering
	\includegraphics[width=.8\linewidth]{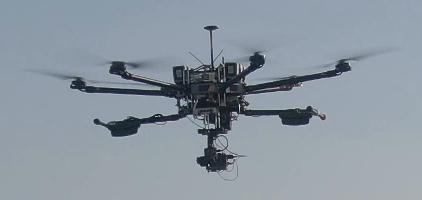}
	\caption{UAV platform used for the field experiments. It is equipped with a multimedia camera on a 3-axis gimbal (bottom).}
	\label{fig:SingleDroneFlying}
\end{figure}

The aerial platforms used in our experiments (see Fig.~\ref{fig:SingleDroneFlying}) were three custom-designed hexacopters made of carbon fiber with a size of $1.80 \times 1.80 \times 0.70~m$, $13~kg$ of maximum take-off weight and a maximum flight time of $15$ minutes. Each UAV is equipped with: an audiovisual camera mounted on a 3-axis gimbal for stabilization; a RTK GPS receiver; a Pixhawk 2.1 autopilot with ArduPilot; an Intel NUC for running onboard navigation algorithms; a Nvidia TX2 to manage video streaming; and a communication module that uses both LTE and Wi-Fi technology to communicate with the modules on the ground and for inter-UAV communication, respectively.

\subsection{Experimental results}

\begin{figure}[tb]
	\centering
	\includegraphics[width=\linewidth]{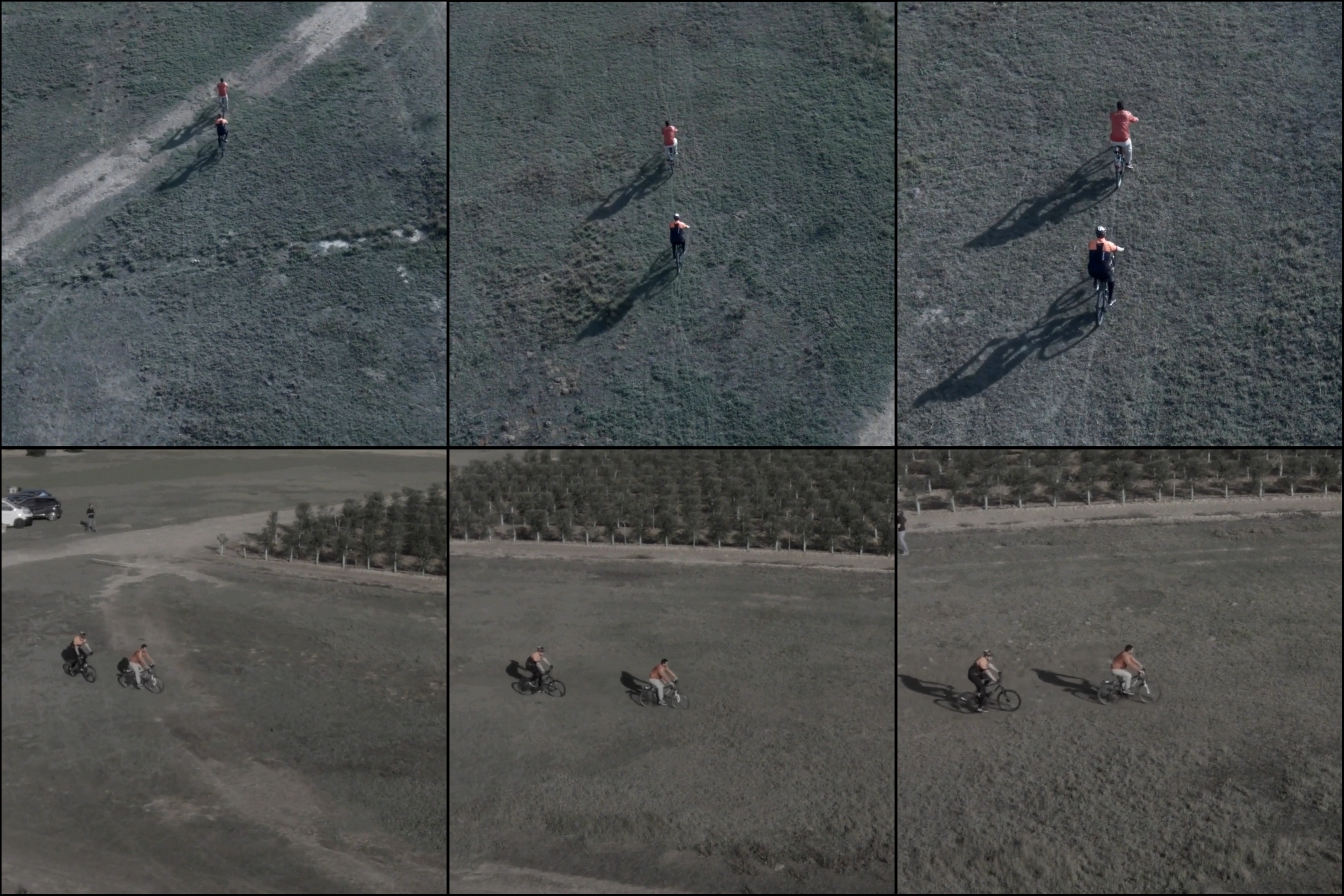}
	\put(-215, 87){\textcolor{white}{T=4s}}
	\put(-136, 87){\textcolor{white}{T=14s}}
	\put( -54, 87){\textcolor{white}{T=35s}}
	\put(-215, 5){\textcolor{white}{T=4s}}
	\put(-136, 5){\textcolor{white}{T=14s}}
	\put( -54, 5){\textcolor{white}{T=35s}}
	\caption{Snapshots of the videos taken onboard during the Establish shot (top row, UAV 2), and the Flyby shot (bottom row, UAV 3). The relative movement of the camera with respect to the targets can be appreciated.}
	\label{fig:TimeComposition}
\end{figure}
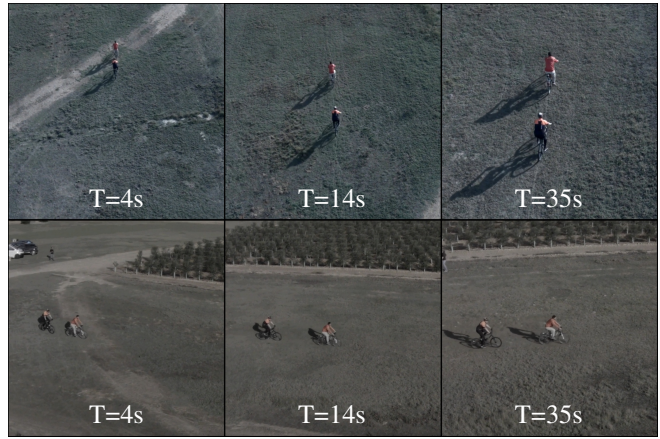

We evaluated the whole system for shooting mission planning and execution during several days in an airfield located close to Seville (see Fig.~\ref{fig:MissionOf3Drones}). We created there a mock-up scenario for outdoor media production, including actual cyclists. We tested the system with multiple missions defined by the Director, considering different types of autonomous aerial shots. In this paper, we present results for two representative experiments that showcase the versatility of our planning algorithm. Planning is done on a ground station before the mission starts, taking as input the filming tasks and the current position and battery level of the UAVs. Even though our architecture allows for re-planning during mission execution in case of unexpected events, this was not the case in the included experiments.
Moreover, our complete system can run visual tracking of the targets on the videostreams to point the 3-axis camera gimbal. In these experiments, as we were interested in the planning part, the cyclists carried a GPS receiver to communicate their position to the UAVs and ease camera tracking. In order to provide a reliable target estimation at high rate, a Kalman Filter was run on board each UAV to integrate measurements coming from the target, and hence, deal with communication latency and lower rates from GPS receivers.

\begin{figure}[tbh]
	\centering
	\includegraphics[width=\linewidth]{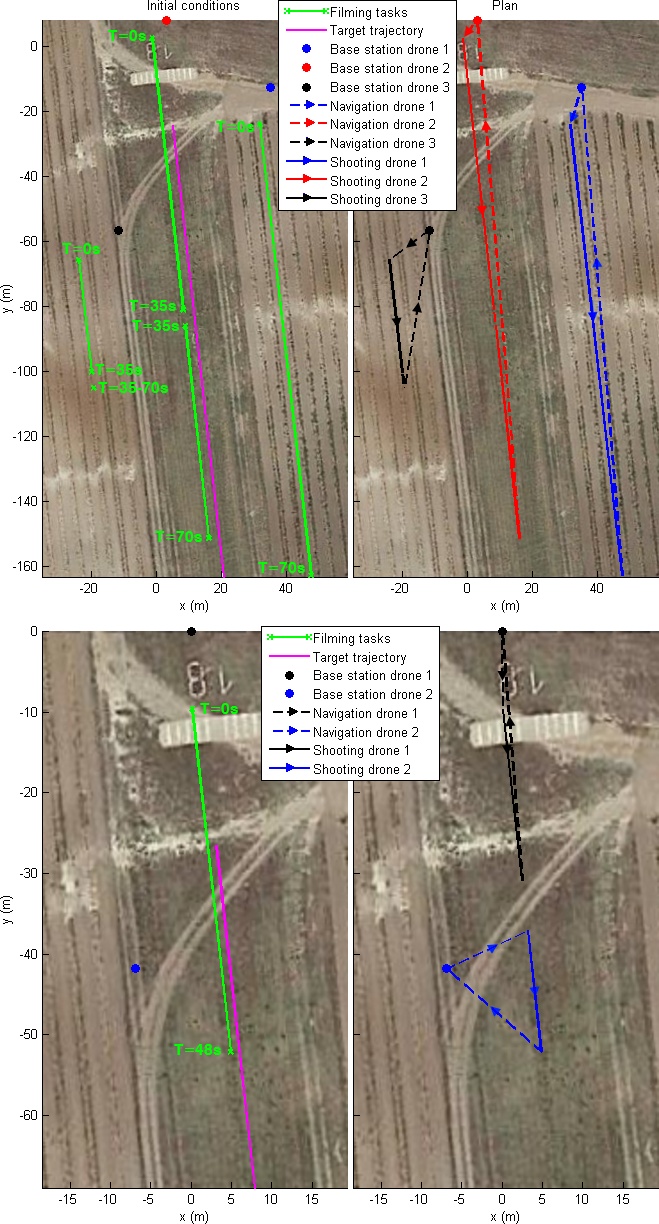}
	\caption{Filming tasks (left graphs) and plans computed for each UAV (right graphs). Plans for each UAV in a different color, with dashed lines indicating UAV navigation and solid lines UAV shooting. Top, first experiment with 3 UAVs. Bottom, second experiment with 2 UAVs.
	}
	\label{fig:MissionsGraph}
\end{figure}

The first experiment consists of a shooting mission involving 3 UAVs filming two actual cyclists. The Director specifies 3 sequences of shooting actions in parallel, with the same starting time and duration (70 seconds) per sequence. The first sequence only includes one shot of type Lateral (UAV follows the target laterally at its same speed). The second sequence has two consecutive filming tasks, an Establish (UAV approaches the target from behind coming closer in distance and height) and a Chase (UAV follows the target from behind at a constant distance), respectively. The third sequence has also two consecutive tasks, a Flyby (UAV starts behind the target at a lateral distance but it catches up with it to overcome it) and a Static (UAV stays still filming a panoramic view). Figure~\ref{fig:TimeComposition} shows a composition of images taken from the UAV cameras during some of the parallel shots are depicted~\footnote{A video of the complete experiment can be seen at \href{https://youtu.be/nRM-TJ2njtg}{https://youtu.be/nRM-TJ2njtg}}.  
The target trajectories and the plan computed can be seen in Fig.~\ref{fig:MissionsGraph} (top). As there are 3 available UAVs with enough battery to cover the mission, the HLP computes a plan assigning directly one sequence of shooting actions to each UAV. We used a computer with an Intel i7-6700HQ@2.6GHz and 16 GB RAM as ground station to run the HLP module, obtaining the plan in $1.038~ms$. All tasks were fully covered by the computed plan ($CR=1$). 

The second experiment consists of a shooting mission involving 2 UAVs filming one cyclist. The Director specifies a single Chase shot ($48~s$) but too long to be covered by the same UAV due to battery limits (we set up shorter battery endurance in the UAVs to enforce this situation). Therefore, the computed plan (in $0.817~ms$) assigns one part of the Chase to each of the available UAVs. The plan computed can be seen in Fig.~\ref{fig:MissionsGraph} (bottom)~\footnote{A video of the complete experiment can be seen at \href{https://youtu.be/-8Y8OGbHE9c}{https://youtu.be/-8Y8OGbHE9c}}. The first UAV starts covering the task from the beginning, and then returns to the Base Station after running out of battery. The second UAV replaces the first one at a relay point and covers the remaining part of the task. Due to safety reasons, both UAVs do not arrive in the relay point at the same time instant, so there is a short time interval in between the end of the first filming and the start of the second. 
Note that the UAV autopilots had a speed limitation during the experiment for safety reasons, as they had to fly near the cyclists. Due to that, we recreate the cycling race at a lower speed than usual, so that UAVs were able to overcome the cyclists.  
In particular, 8 seconds of the original shooting action were not cover during the relay operation ($CR=0.833$). We ran several missions obtaining similar results, but we did not include them here due to space limitations. 

\section{Conclusions}
\label{sec:conclusions}

In this paper, we presented an algorithm for planning cinematography missions with multiple UAVs covering outdoor events. The strategy is based on an efficient dynamic programming approach for one UAV, that is used in an iterative way to produce an approximate solution for the multi-UAV problem. Although the algorithm is deterministic and assumes a motion model to predict targets' trajectories, our final architecture allows us to monitor mission execution and re-compute new plans online in case of contingencies like UAV failures or deviations from the original plan.

Results in simulation proved that using a number of UAVs equal to the number of overlapping tasks, 80\% of the total filming time can be covered, which somehow justify our assumption of small teams. We also demonstrated empirically that our approximate solutions are close to the optimum. Moreover, we present field experiments to show the applicability of our approach for media production in realistic cycling race scenarios.

Future work will focus on considering a constrained model that minimizes the number of UAVs per task. It is reasonable to impose that a UAV leaves a task only if it is running out of battery or the end of task is reached. Thus, we try to reduce the complexity of the optimization problem pursuing exact optimal solutions for the multi-UAV case.

\balance 

\section*{Acknowledgment}
This project has received funding from the Spanish Ministry of Economy and Competitiveness project (MTM2016-76272-R AEI/FEDER,UE), the European Union's Horizon 2020 research and innovation programme under grant agreement No 731667 (MULTIDRONE), the Marie Sk\l{}odowska-Curie grant agreement No 734922 and the MULTICOP project (Junta de Andalucia, FEDER Programme, US-1265072).

\bibliographystyle{IEEEbib}
\bibliography{bibliography}

\end{document}